\pdfoutput=1

\documentclass[11pt]{article}

\usepackage{acl}

\usepackage{times}
\usepackage{latexsym}

\usepackage[T1]{fontenc}

\usepackage[utf8]{inputenc}

\usepackage{microtype}

\usepackage{inconsolata}

\usepackage{graphicx}

\usepackage{tabularx}
\usepackage{booktabs}
\usepackage{amssymb}
\usepackage{multirow}
\usepackage{subcaption}
\usepackage{tcolorbox}
\newtcolorbox{mybox}[3][]
{
  colframe = #2!25,
  colback  = #2!10,
  coltitle = black,  
  title    = {#3},
  #1,
}

\newcommand{\datain}{$D^P$}
\newcommand{\dataout}{$D^A$}
\newcommand{\dataoutselect}{$D^{A\_sim}$}
\newcommand{\dataoutrandom}{$D^{A\_rand}$}
\newcommand{\dataoutselectall}{$D^{A\_simall}$}
\newcommand{\dataoutrandomall}{$D^{A\_randall}$}
\newcommand{\datainsame}{$D^{P\_same}$}
\newcommand{\datainswap}{$D^{P\_swap}$}
\newcommand{\ourbert}{CTI-BERT}

\title{Cyber-Attack Technique Classification \\ Using Two-Stage Trained Large Language Models} 

\author{  
Weiqiu You \\
University of Pennsylvania \\
  \texttt{weiqiuy@seas.upenn.edu} \\\And
  Youngja Park \\
  IBM T.J. Watson Research Center\\
  \texttt{young\_park@us.ibm.com} \\}

\begin{document}
\maketitle

\begin{abstract}
Understanding the attack patterns associated with a cyberattack is crucial for comprehending the attacker's behaviors and implementing the right mitigation measures.
However, majority of the information regarding new attacks is typically presented in unstructured text, posing significant challenges for security analysts in collecting necessary information.

In this paper, we present a sentence classification system that can identify the attack techniques described in natural language sentences from cyber threat intelligence (CTI) reports.
We propose a new method for utilizing auxiliary data with the same labels to improve classification for the low-resource cyberattack classification task.
The system first trains the model using the augmented training data and then trains more using only the primary data.

We validate our model using the TRAM data\footnote{https://github.com/mitre-attack/tram} and the MITRE ATT\&CK framework.
Experiments show that our method enhances Macro-F1 by 5 to 9 percentage points and keeps Micro-F1 scores competitive when compared to the baseline performance on the TRAM dataset.
\end{abstract}

\section{Introduction}
The rapid growth of cyberattacks, both in numbers and techniques, presents significant challenges to companies, often leading to incidents of data theft, financial losses, and disruptions to critical infrastructure.
To promptly respond to the cyber threats, it is vital for security analysts to collect and process comprehensive information on the threats, including the tactics, techniques and procedures (TTPs) employed in the attacks.

However, much of the information on new cyber attacks appear in unstructured documents, such as blogs, news articles and tweets. 
Although these cyber-threat intelligence (CTI) reports can provide valuable insights into the on-going attacks and the evolving threat landscape, 
collecting relevant information from a large volume of unstructured reports is very time consuming and labor intensive.
For instance, threat hunters must sift through several lengthy documents to understand the attacker's TTPs before they can create detection rules and  respond effectively to the attack.

\begin{table*}[ht]
\begin{small}
\begin{tabular}{@{}p{0.15\linewidth}p{0.45\linewidth}p{0.35\linewidth}@{}}
\toprule
Technique & TRAM (CTI Report)  & MITRE (Description) \\ \midrule
\multirow{7}{*}{Abuse} \multirow{7}{*}{Elevation} \multirow{7}{*}{Control} \multirow{7}{*}{Mechanism}& \multirow{3}{*}{sudo}  & Adversaries may circumvent mechanisms designed to control elevate privileges to gain higher-level permissions.                                  \\\cmidrule{2-3}
& (To bypass UAC) configurable setting for the process to abuse Other than these, new coding algorithm has been introduced.  
& Most modern systems contain native elevation control mechanisms that are intended to limit privileges that a user can perform on a machine.                            
\\\midrule
\multirow{7}{*}{Access Token} \multirow{7}{*}{Manipulation}         & The tokens for each platform are hardcoded within the sample:November 2016 to January 2017: "Evil New Year" CampaignIn the early part of 2017, Group123 started the "Evil New Year" campaign. 
& Adversaries may modify access tokens to operate under a different user or system security context to perform actions and bypass access controls.                       
\\\cmidrule{2-3}
& The Trojan uses the access token to write the string above to the first file uploaded to Google drive whose filename is .txt. 
& A user can manipulate access tokens to make a running process appear as though it is the child of a different process or belongs to someone other than the user that started the process. 
\\\midrule
\multirow{10}{*}{Access} \multirow{10}{*}{Discovery} 
& This may include information about the currently logged in user, the hostname, network configuration data, active connections, process information, local and domain administrator accounts, an enumeration of user directories, and other data.
& Adversaries may attempt to get a listing of accounts on a system or within an environment.                        \\\cmidrule{2-3}
& The PowerShell script collects all possible information on the user and the network, including snapshots, computer and user names, emails from registry, tasks in task scheduler, system information, AVs registered in the system, privileges, domain and workgroup information.      & This information can help adversaries determine which accounts exist to aid in follow-on behavior.               \\ \bottomrule
\end{tabular}
\caption{Example sentence from CTI data and MITRE ATT\&CK data. CTI sentences are extracted from the the TRAM dataset and contain named entities of specific files and tokens. MITRE data, on the other hand, are general descriptions of attack techniques, so they are higher level statements of attack techniques' behaviors.}
\end{small}
\label{tab:example}
\end{table*}

Recently, there have been efforts to automatically extract cybersecurity attack techniques (TTPs) from CTI reports~\cite{TTPDrill,AttacKG,rcATT,Sauerwein,raid}.
These existing tools assign TTP labels to either  IoCs~\cite{TTPDrill,AttacKG}, short phrases~\cite{raid}, or entire documents~\cite{rcATT,Sauerwein}. 
These approaches have limitations. 
The document-level TTP classification  does not provide actionable insights, e.g., how exactly a TTP was used. Security analysts still must read the documents and find the TTP-related information manually.
The IoC or phrase-level TTP classification  provides partial knowledge and lacks the contextual information. 
Further, these methods typically employ a pipeline system consisting of several NLP (natural language processing) and machine learning models. 
However, the capabilities of state-of-the-art NLP technologies for cybersecurity text are still limited.
In particular, pipeline-based approaches can propagate errors where mistakes from the previous steps negatively impact the performance of subsequent steps. 
A machine learning model that is trained end-to-end can address the error propagation problem.

In this work, we present a novel classification model that can classify natural language sentences into their respective TTPs from the MITRE ATT\&CK framework. 
It is worth noting that generating a training dataset with a sufficient number of CTI sentences labeled with TTPs is a highly challenging task.  
We propose to solve the data sparseness in two ways. 

Firstly, we utilize the TTP descriptions from MITRE ATT\&CK as an additional training data. 
MITRE ATT\&CK is a knowledge base that contains descriptions of attack tactics, techniques, mitigation, software, etc.
For each attack technique, there is a corresponding description of it in the MITRE knowledge base.
We use these descriptions of attack techniques as additional data to train the attack technique classification.
However, the style and vocabulary of CTI sentences are very different from that of the descriptions in MITRE ATT\&CK, as shown in Table~\ref{tab:example}.
In this case, simply adding the MITRE data to CTI data does not improve the classification performance.
To address this problem, we introduce a similarity-based data augmentation and a novel two-stage training method. 
We first train the classification model using the combined data and then continue training the model longer using only the CTI data.
This two-stage training allows the model to benefit from the additional data while minimizing the negative effect of the out-of-distribution data.

Secondly, we base our classification models on a large language model (LLM) pretrained with a collection of CTI documents.  
In recent years, LLMs, such as BERT~\cite{Devlin2019BERTPO}, have shown a superior performance for many NLP applications.
These foundation models are first trained on a large amount of unlabeled text to gain general semantic knowledge, and then fine-tuned on a relatively small amount of labeled data to optimize for a specific task, such as attack technique classification. 
We train a BERT model (called \ourbert{} hereafter) using a collection of CTI reports, vulnerability descriptions and security-related academic publications. 
Our system leverages the LLM's ability to capture the contextual meaning and produces a higher accuracy than baseline methods.

Our main contributions are the following:
\begin{enumerate}
    \item We propose a similarity-based two-stage training pipeline to use selected auxiliary data to help training with primary data.
    \item Our experiments validate the effectiveness of our method on cyberattack classification, showing that it improves Macro-F1 scores significantly while maintaining competitive Micro-F1 scores.
    \item We demonstrate how domain-specific LLMs are very effective in tackling the data sparseness problem.
\end{enumerate}


\begin{figure*}[ht]
\centering
\begin{subfigure}{\textwidth}
\centering
   \includegraphics{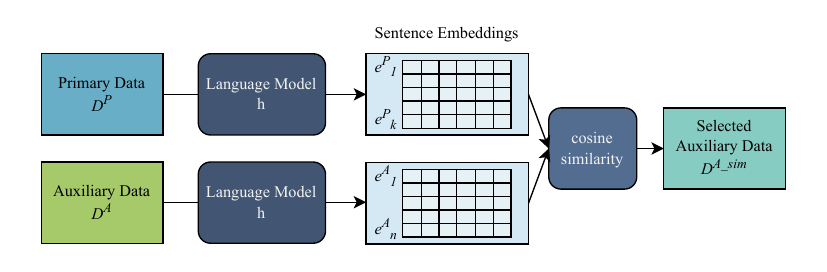}   
   \caption{Cosine-similarity-based Auxiliary Data Selection}
   \label{fig:data_selection}
\end{subfigure}
\begin{subfigure}{\textwidth}
\centering
\vspace{1em}
   \includegraphics{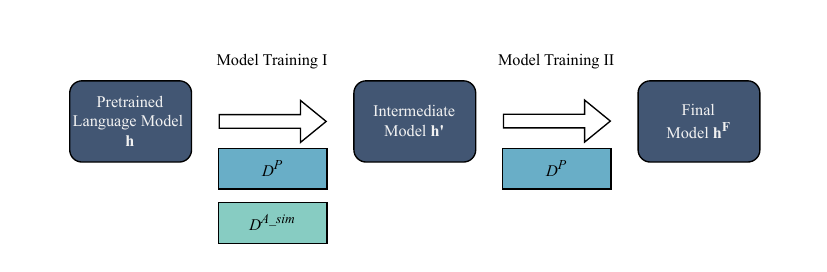}     
   \caption{Two-stage Model Training}
   \label{fig:pattern-feature}
\end{subfigure}
\caption{System Overview. For the given primary data (\datain) and auxiliary data (\dataout), we first select a subset of the auxiliary samples (\dataoutselect) based on their similarity with \datain. The model training is done using \datain and \dataoutselect at the first stage and then with only \dataoutselect at the second stage.}
\label{fig:method}
\end{figure*}

\section{Two-stage Training with LLMs}
We aim to build a generic framework for the \textit{low-resourced} and \textit{imbalanced} attack technique classification task.
We solve the problem by utilizing a small amount of external data.
Given a small primary dataset, \datain, and an auxiliary dataset, \dataout,
our objective is to improve the model's classification performance on \datain. 
We assume that both primary and auxiliary datasets belong to the same domain with the same label sets. 
However, the writing style and vocabulary in the two datasets differ significantly, and, thus, simply adding the auxiliary data to the primary data often deteriorates the classification results.

We propose a two-stage model training technique with similarity-based auxiliary data selection to address the low-resource class-imbalance problem.
Figure~\ref{fig:method} shows the high-level architecture of our approach. 
We first select part of \dataout\ that is similar to \datain\ and augment the training data with the selected subset of \dataout. 
Then, we fine-tune a pretrained language model using the augmented data first and, then, fine-tune the model further in the second stage only  with \datain\ to steer the final model closer to \datain.

\subsection{Similarity-based Auxiliary Data Incorporation on Minority Classes}
\label{sec:similarity-augmentation}
The simplest way to use auxiliary data would be to concatenate the two datasets \datain\ and \dataout.  
However, due to the distribution shift, adding auxiliary data can harm the model's performance, especially for the majority classes if the classes already have some high quality data. 
To minimize the negative effect of distribution shift, we augment \datain\ only for minority classes by adding  a subset of \dataout\ that is more similar to \datain 
to training. In this work, we define a minority class as one having fewer than $k$ samples.

There are different methods to assess how similar two sentences are, including minimum-edit distance~\citep{Levenshtein1965BinaryCC}, cosine similarity~\citep{Salton1975AVS}, dot-product~\citep{Salton1975AVS}, etc.
With LLMs, the most widely used method to measure the sentence similarity is to convert the sentences into their embeddings (i.e., vector representations) and calculate the cosine similarity between the sentence embeddings.

We use the embedding-based cosine similarity. To select similar samples, we use a pretrained language model, BERT~\cite{devlin-etal-2019-bert}, to compute the pairwise cosine similarity between $D^A_c$ and $D^P_c$, which represent the auxiliary and primary data for a class $c$ respectively. 
For each sentence $s^A_i \in  D^A_c$, its similarity to $D^P_c$ is defined as:
\begin{equation} 
\mathrm{Sim}(s^A_i, D^P_c) = \max_{s^P_j \in D^P_c} \cos (\mathbf{h}(s^A_i) , \mathbf{h}(s^P_j))
\end{equation}
 where  $\cos(\mathbf{h}_1, \mathbf{h}_2)$=$\frac{\mathbf{h}_1^{\intercal} \mathbf{h}_2}{||\mathbf{h}_1||\cdot||\mathbf{h}_2||}$, 
and $\mathbf{h}(s^A_i)$ is the embedding of $s^A_i$. 
Then, we take the most similar $k - |D^P_c|$ sentences from $D^A_c$ to augment $D^P_c$. 
This way, we make each minority class have up to $k$ sentences. We denote the set of sentences selected based on the similarity to the \datain\ as \dataoutselect.

\subsection{Domain-specific Language Model}
Although training on general-domain text data allows BERT to learn the English language, it has some limitations when we process text in a specific domain,  such as law, medical, and security, etc. due to their own specific vocabulary. 
To improve the performance of security text understanding, we use a BERT model trained from scratch for the cybersecurity domain using a high quality security text dataset~\cite{park-you-2023-pretrained}. 
The pretraining dataset contains about 1.2 billion words collected from various cybersecurity threat intelligence (CTI) reports covering key security topics such as security campaigns, malware, threat actors and vulnerabilities.
The dataset includes news articles, blogs and APTnotes\footnote{https://github.com/aptnotes/data},  MITRE datasets on attacks\footnote{https://attack.mitre.org/, https://capec.mitre.org/} and vulnerabilities\footnote{https://cve.mitre.org/,  https://cwe.mitre.org/}, academic publications from security conferences, security textbooks, Wikipedia pages belonging to the ``Computer Security'' category.

%


\subsection{Two-stage Model Training}
\label{sec:two-stage}
As discussed earlier, simply augmenting the training data with auxiliary data can deteriorate the performance (see our experiment results in Table~\ref{tab:main_results}).
In order to gain back the loss in performance due to the data distribution shift, we propose a two-stage training, where we first train a model using both  \datain\  and  \dataoutselect\ and then train the model more on \datain\ only.
The second stage training can bring the model closer to \datain.

\section{Experiments}
We validate our method focusing on the cyber-attack technique classification task, i.e., classifying CTI sentences to their attack technique types.

\subsection{Data}

We use the TRAM\footnote{https://github.com/center-for-threat-informed-defense/tram} and MITRE ATT\&CK\footnote{https://attack.mitre.org} datasets as primary and auxiliary data respectively.
The MITRE dataset contains the description for each attack technique. 
The TRAM dataset contains CTI sentences extracted from security news and technical reports. The sentences are  labeled with the attack techniques from MITRE ATT\&CK. 

\paragraph{\textbf{Data Splits and Preprocessing}}
The original TRAM dataset has a few limitations. 
First, it only has the train and test splits (no validation split), and the train and test sets are created by duplicating all examples four times and randomly splitting into a ratio of 4:1, resulting in many examples appearing in both splits. 
Further, MITRE ATT\&CK organizes attack techniques by different targets--Enterprise, Mobile and Industrial Control Systems, 
and some techniques can apply to multiple targets such as enterprise and mobile. 
In TRAM, if an attack technique is used for different targets, they are labeled as different classes. 

For our experiments, we remove all duplicates from the original train and test splits and combine the classes with the same name, regardless of the target systems. 
Further, to evaluate the model over all the classes, we remove the classes with fewer than three examples to make sure that each class has at least one example in the splits. 
After these preprocessing, we obtain 1,491 TRAM sentences with 73 classes out of the original 264 classes. We then split the sentences into the train, development and test sets in a ratio of 2:1:1. 
Finally, we collect the descriptions for the 73 classes from MITRE, resulting in 2,637 sentences. 
We use the pre-processed TRAM data as \datain{} and the MITRE descriptions as \dataout.

\paragraph{\textbf{Data Statistics}}
Table~\ref{tab:tram-data} shows the summary of our datasets.
As we can see, the TRAM data has the following characteristics: 1) \textit{Low-resource}: only 754 training examples. 2) \textit{Unbalanced}: 14 classes have only one sentence in the training split. 3) \textit{Many-classes}: 73 classes for this small amount of training data.
\begin{table}[th]
    \centering
    \begin{small}
    \begin{tabular}{ccccc}
    \toprule
         Data Set & \datain\ Train & \datain\ Dev & \datain\ Test & \dataout \\\midrule
         \# Sent & 754 & 355 & 382 & 2,637 \\
    \bottomrule
    \end{tabular}
    \end{small}
    \caption{Summary of our datasets: 73 classes, 1,491 primary sentences and 2,637 auxiliary sentences.}
    \label{tab:tram-data}
\end{table}

\begin{figure}[t]
\includegraphics{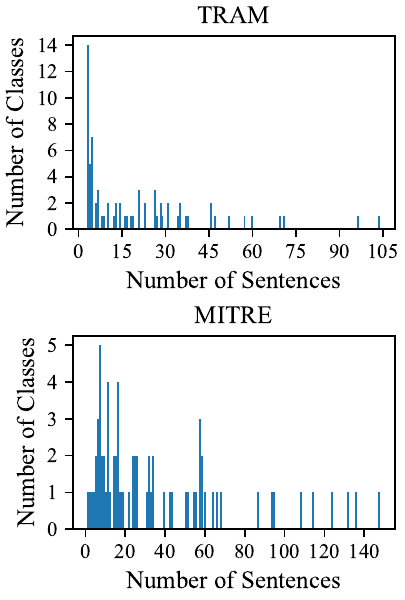}
\centering
\caption{Distribution of classes based on the number of member sentences. The TRAM data is highly imbalanced, having 14 classes with only 3 sentences (one each in train, validation and test splits), while the MITRE data is more balanced, with only 3 classes with less than 3 sentences.}
\label{fig:data-histogram}
\end{figure}

Figure~\ref{fig:data-histogram} shows the distribution of class sizes in MITRE and TRAM. We can see that the MITRE dataset has a more balanced distribution with few minority classes, while TRAM has a large number of classes with only one or a few sentences. Having only a few sentences makes it hard for the model to learn to generalize, and thus adding additional sentences from MITRE can help significantly with the minority classes.

\begin{table*}[t]
\setlength\tabcolsep{3.3pt}
\begin{tabular}{lllllllll}
\toprule
                        \multirow{2}{*}{Models}& \multicolumn{2}{c}{BERT}  & \multicolumn{2}{c}{SecBERT} & \multicolumn{2}{c}{SecureBERT} & \multicolumn{2}{c}{\ourbert{} (ours) } \\ \cmidrule{2-9}
 & Micro    & Macro      & Micro  & Macro   & Micro   & Macro  & Micro  & Macro   \\ \midrule
            \textit{Primary Data Only} \\
(1) \datain{} \textit{(baseline 1)}             & 61.2              & 38.0             & 63.1         & 41.6     &  63.8 &  43.3    & 69.1 & 47.3    \\
(2) \datain$\rightarrow$\datain &61.1	 & 38.5 & 62.5 & 42.6 & 64.6 & 44.5 & 67.9 & 46.5 \\ \midrule
\textit{Augmentation w/ all of auxiliary} \\
(3) \datain+\dataout{} \textit{(baseline 2)}      & 56.0              & 42.9 & 59.3         & 48.0        &56.0 &  43.1  & 65.7 & 55.1    \\
(4) \datain+\dataout$\rightarrow$\datain  & 61.0              & \textbf{44.9}         & 63.6         & \textbf{50.4}        &  62.9 & 47.4  & 70.9 & \textbf{58.0}  \\ \midrule
\textit{Augmentation w/ similar auxiliary} \\
(5) \datain+\dataoutselect         & 59.4              & 42.4       & 63.5         & 48.3    & 64.0 &   48.7  & 70.1 & 56.8  \\
(6) \datain+\dataoutselect$\rightarrow$\datain \textit{(Proposed)} & \textbf{62.9}           & 44.8    & \textbf{63.8}         & 48.2   &  \textbf{65.4}&   \textbf{50.2}  & \textbf{71.3} & 56.1   \\
\bottomrule
\end{tabular}
\caption{Average Micro-F1 and Macro-F1 scores over five runs for different models. `$\rightarrow$' indicates two-stage training. The average standard deviation is 0.014. The best results are highlighted in bold.}
    \label{tab:main_results}
\end{table*}

\subsection{Training Details}
\label{sec:training}
We fine-tune both the first-stage and second-stage models for the 73-way multiclass classification.
The classification models are trained for 50 epochs for each stage, with the batch size of 16 and the learning rate of 2e-5.
After 1,000 warmup steps, the learning rate is varied according to the formula in \citet{Vaswani2017}.
We use the Adam optimizer with $\beta_1=0.9$, $\beta_2=0.999$, and weight decay of 0.01.
The code is implemented using PyTorch and HuggingFace Transformers\footnote{https://github.com/huggingface/transformers}.

We select $k=10$ as the threshold cutoff for minority classes based on preliminary ablations.
The hyper-parameters are selected base on the validation split of \datain{} (i.e., \datain Dev in Table~\ref{tab:tram-data}).
The experiments are performed on Quadro RTX 6000 or V100 GPUs. 

\subsection{Baseline Systems}
We compare our two-stage classification method with two baseline methods. 
The baseline models are trained once using (1) only the primary data (\datain) and (2)  all of the primary and auxiliary data (\datain+\dataout).  
We also compare 2-stage training with 1-stage training for all cases (primary data only, primary+all auxiliary data, primay+similar auxiliary data), resulting in 6 different cases.

Furthermore, we compare our model with the general domain BERT model (\texttt{bert-base-uncased}) and two other pretrained foundation models for the cybersecurity domain: SecBERT~\cite{secbert} and SecureBERT~\cite{securebert}.
Similarly to \ourbert{}, SecBERT\footnote{https://huggingface.co/jackaduma/SecBERT} is a BERT model pretrained from scratch using APTnotes, Stucco-data, CASIE and data from the SemEval-2018 Task-8.
On the other hand, SecureBERT updates the general-domain RoBERTa model~\cite{roberta} using a security corpus.
We fine-tuned these three baseline models using the same training setup as described in Section~\ref{sec:training}.

\begin{table*}[t]
\begin{tabular}{lllllllll}
\toprule
\multirow{2}{*}{~~~~~Models}& \multicolumn{2}{c}{BERT}  & \multicolumn{2}{c}{SecBERT} & \multicolumn{2}{c}{SecureBERT} & \multicolumn{2}{c}{\ourbert{} (ours)}  \\ \cmidrule{2-9}
 & Micro    & Macro      & Micro  & Macro   & Micro   & Macro  & Micro  & Macro   \\ \midrule
\textit{\textbf{Self Data Augmentation}} \\
(a) \datain+\datainsame$\rightarrow$\datain & 61.6 & 39.2    & 63.5    & 42.7       &  63.1  &   41.8 & 69.1 & 48.4 \\
(b) \datain+\datainswap$\rightarrow$\datain & 61.6 & 39.2   & 63.0     & 42.7        & 63.1 &    41.8  & 69.1 & 48.4  \\
\midrule
\textit{\textbf{Using Auxiliary Data}} \\
(c) \datain+\dataoutrandom$\rightarrow$\datain &  61.6  &   43.4  &   63.4   &  49.1  & \textbf{66.2} &   49.4 & 69.0 & 55.0   \\
(d) \datain+\dataoutselect$\rightarrow$\datain & \textbf{62.9}              & \textbf{44.8}    & \textbf{63.8}         & 48.2   &  65.4 & 50.2  &\textbf{71.3} & 56.1   \\ 
\bottomrule
\end{tabular}
\caption{Ablation Results for Minority Class Augmentation. The models augment only minority classes with fewer than $k$ samples.
\datainsame\ oversamples sentences from the same classes ; \datainswap\ randomly swaps tokens within each oversampled sentence in \datainsame; \dataoutrandom\ randomly selects examples from \dataout; \dataoutselect\ selects most similar examples from \dataout. 
}
\label{tab:ablation_results_minority}
\end{table*}

\begin{table*}[t]
\begin{tabular}{lllllllll}
\toprule
\multirow{2}{*}{~~~~~Models}& \multicolumn{2}{c}{BERT}  & \multicolumn{2}{c}{SecBERT} & \multicolumn{2}{c}{SecureBERT} & \multicolumn{2}{c}{\ourbert{} (ours)}  \\ \cmidrule{2-9}
 & Micro    & Macro      & Micro  & Macro   & Micro   & Macro  & Micro  & Macro   \\ \midrule
(e) \datain+\dataoutrandomall$\rightarrow$\datain &  60.9  &   43.6 &   62.8  & \textbf{49.4} &  65.3  &   \textbf{50.5} & 69.6 & \textbf{56.8}   \\
(f) \datain+\dataoutselectall$\rightarrow$\datain &       61.5        &  43.8  &  62.4     &  48.1  &  64.6 &  47.5  & 68.6 & 53.2   \\
\bottomrule
\end{tabular}
\caption{Ablation Results for All Class Augmentation. The models augment all classes.
 \dataoutrandomall\ randomly selects $k$ examples from each class in \dataout; \dataoutselectall\ selects top $k$ similar examples from each class in \dataout. ($k=10$)
 }
\label{tab:ablation_results_all}
\end{table*}

\section{Results and Discussion}

Table~\ref{tab:main_results} summarizes the results.
Note that, for the \datain+\dataoutselect\ setting, we augment only minority classes, which have fewer than 10 sentences, so that all classes have at least 10 sentences.
For each case, we train five models with five different seeds and report the average Micro and Macro-F1 scores over the 5 runs. 
Micro-F1 shows how accurate the model is at predicting for all sentences, and Macro-F1 shows how accurate the model is at predicting for all classes.

We also perform several ablation studies with different sampling techniques--oversampling of primary data and  random sampling from auxiliary data. Further, we compare cases where the auxiliary data is incorporated for only minority classes vs. all classes. The details and the results of the ablations are shown in Table~\ref{tab:ablation_results_minority} and \ref{tab:ablation_results_all} respectively.  From these experimental results, we obtain the following findings.

\subsection{Auxiliary Data}
To evaluate whether adding auxiliary MITRE data helps with prediction on the primary TRAM data, we train \datain+\dataout{} model ((3) in Table~\ref{tab:main_results}).
We produce the \datain+\dataout{} model by simply training with the combined primary and auxiliary data.
The intuition is that, if auxiliary data is in the same distribution as the primary data, then simply adding it should already help. 

We can see from Table~\ref{tab:main_results} that augmenting TRAM with MITRE (\datain+\dataout) increases Macro-F1 but lowers Micro-F1 compared to the results of \datain.
MITRE has more evenly distributed data across classes as shown in Figure~\ref{fig:data-histogram}.
The increased Macro-F1 shows that balanced data helps more classes to have better performance.
However, the fact that Micro-F1 lowers shows that there is a gap between the style of \datain{} and \dataout{}.
This aligns with how \datain{} contains sentences from CTI reports, while \dataout{} contains sentences from descriptions.

\subsection{Two-stage Training}
Since simply adding auxiliary MITRE data does not help, we test the models (4) (\datain+\dataout$\rightarrow$\datain) in Table~\ref{tab:main_results} to evaluate if continuing training on primary data on the model (3) helps.
The intuition is that, continuing training on only the primary data might steer the model back to the distribution of primary data, while keeping the benefit of auxiliary data on rare classes.

We can see from Table~\ref{tab:main_results} that models (4), compared to models (3), not only gains back Micro-F1, but also improves more on Macro-F1.
This shows the usefulness of our two-stage training pipeline in resolving the mismatch between distributions of TRAM and MITRE.

Moreover, we do a sanity check of whether two-stage training only benefits from training the model longer, instead of from auxiliary data.
By comparing (4) and (2) in Table~\ref{tab:main_results}, we can see that simply training the model with primary data for longer does not help with Macro-F1.

\subsection{Adding Similar Data to Rare Classes}
Since adding MITRE data helps with rare but not common classes, we ask the question: Can we take advantage of this and only use it for rare classes?
We add MITRE sentences to these rare classes such that they have up to $k$ sentences, with rare classes defined in Section~\ref{sec:training}.
In our preliminary experiments, we have tried $k=5, 10, 15$ as the threshold, and $k=10$ produced the best Micro and Macro F1 scores on the dev. split, so we use $k=10$ for all our experiments.

Another question lies in how to select the sentences from MITRE to add.
A natural way is to add the sentences that are most similar to TRAM sentences, such that there is less distribution shift.
We use the cosine-similarity-based selection method described in Section~\ref{sec:similarity-augmentation} to select the most similar sentences.

We can see from Table~\ref{tab:main_results} that models (6) consistently outperforms (4) in Micro-F1 for all models, while having competitive Macro-F1.
Even for the one-stage versions, models (5) also outperforms (3) in Micro-F1 and sometimes Macro-F1.
This shows that selecting more similar data to add to only rare classes helps the best.

\subsection{Rare Classes Only}
However, is the performance improvement from adding similar data, or from adding data to rare classes, or both?
To evaluate if adding auxiliary data only to rare classes is better than adding to all classes, we have ablations of models (c) from Table~\ref{tab:ablation_results_minority} vs. models (e) from Table~\ref{tab:ablation_results_all} and models (d) from Table~\ref{tab:ablation_results_minority} vs. models (f) from Table~\ref{tab:ablation_results_all}.
Here, when we add the examples to all classes, we also only add up to $k=10$ sentences to compare with the rare class class.

We can see from the comparison between (c) and (e) that using randomly selected auxiliary data for only rare classes is better than adding them to all classes.
Similarly, the comparison between (d) and (f) shows that adding similar auxiliary data for only rare classes is also better than for all classes.

\subsection{Similar vs. Random}
To evaluate if selecting similar data is necessary, we compare with adding randomly selected data from auxiliary data.
The intuition is that, if adding randomly selected data helps more than adding data selected by similarity, then there is no need to select auxiliary data based on similarity.

We can see from (d) vs. (c) in Table~\ref{tab:ablation_results_minority} and (f) vs. (e) in Table~\ref{tab:ablation_results_all} that selecting similar data helps Micro and Macro F1s for BERT, but not necessarily for SecBERT, SecureBERT and \ourbert{}.
This result is surprising to us, but one possible explanation is that security domain foundation models already have good representations for the security domain text data.
Therefore, they are able to use random auxiliary data with no worse effectiveness as similar data.
Yet, for a general-domain model such as BERT, it does not have a good understanding of how sentences in CTI reports relate to the descriptions of attack techniques.
It then need more supervision from auxiliary data that are more similar to the primary data.

\subsection{Auxiliary Data vs. Oversampled Primary Data} 
Lastly, to confirm if auxiliary data adds new information to primary data, we compare it with simply oversampling the primary data (see Table~\ref{tab:ablation_results_minority}).
We compare with two oversampling methods: (a) repeating sentences and (b) repeating sentences but with some words swapped within each sentence.
For the auxiliary data, we use both (c) randomly selected auxiliary data and (d) similarity-based selected auxiliary data.
All experiments  augment data only to minority classes.
The intuition is that, if auxiliary data has different information that is useful for the primary data, then adding it should produce a better performance than simple data augmentation techniques on just the primary data.

From Table~\ref{tab:ablation_results_minority}, we can see that both adding similarity-based and adding randomly selected auxiliary data boost both Micro and Macro-F1 much more than data augmentation on primary data alone.
This confirms that we can often benefit from auxiliary data even if they are not in the same style as primary data.

\section{Related Work}
\subsection*{Attack Technique Classification}
Prior work on security information extraction from threat reports has primarily focused on IoCs and vulnerabilities~\cite{tudor,IoCMiner,twiti,park}.
Recently, several works have proposed various methods to automatically extract TTP-related knowledge from unstructured text.

\citet{threat_report2022, Bridges2017} summarize datasets and methods in CTI extraction.
\citet{Tounsi2017} define four CTI categories and discuss existing CTI tools and research trends.
\citet{Tuma2018ThreatAO} review 26 cyberthreat analysis methodologies.

 \citeauthor{TTPDrill} extracts threat actions (i.e., TTP) from the SVO (Subject-Verb-Object) dependency structure in sentences, where the subject is a malware
instance. The tool considers the object as a TTP  \cite{TTPDrill}. 
\citeauthor{AttacKG} present \textit{AttacKG} which constructs an attack graph by extracting attack-related entities and entity dependencies from CTI reports. 
They also construct a technique template for each attack technique in MITRE ATT\&CK, which is initialized by attack graphs built upon technique procedure examples from the MITRE ATT\&CK knowledge base. 
The techniques used in an attack is determined by aligning the technique templates with the attack graph built from CTI reports~\cite{AttacKG}.  
\citeauthor{rcATT} and \citeauthor{Sauerwein} explore natural language processing (NLP) and machine learning (ML) methods for automated attack technique classification. These works take the reference documents listed in ATT\&CK techniques as the labeled training data and perform a document-level classification, i.e., assigning TTP tags to the entire documents~\cite{rcATT,Sauerwein}.
\citeauthor{raid} treats the attack technique extraction as a sequence tagging task, where the classification model predicts if each token in the sentence belongs to an attack pattern. Then, the system takes each contiguous block of attack pattern tokens as a TTP~\cite{raid}.

The difference between ours and the above work is that these work focus on either a document-level or IoC-level attack technique extraction, while we focus on sentence-level classification.
Sentence-level classification is essential for fine-grained analysis of documents without losing context information.






\subsection*{Transfer Learning with Auxiliary Data}
Transfer learning is commonly done using unlabeled data for pretraining.
ULMFiT~\citep{howard-ruder-2018-universal} demonstrates the effectiveness of finetuning pretrained foundation models on downstream NLP tasks.
BERT~\citep{devlin-etal-2019-bert} shows that pretraining on large corpora of text using masked language modeling and next sentence prediction helps with downstream tasks that are finetuned later.

When we do have labeled auxiliary data for transfer learning, we can make better use of them with specific training pipelines.
\citet{li2020xmixup} proposes to use auxiliary data by mixing of auxiliary data with target data with XMixup (Cross-domain Mixup) in computer vision.
Mixup of data consist of blending two images together, which is not as easy to directly use in the text modality.
\citet{kung-etal-2021-efficient} is the most similar us and they also use a two-stage training pipeline.
They first select data samples from auxiliary tasks based on task similarity from pretrained MT-DNN (Multi-Task Deep Neural Networks), then train MT-DNN model using the selected samples from auxiliary tasks, and then finetune the model on the primary task.
We, on the other hand, show that using auxiliary data selected from cosine similarity with primary data of the same class helps for generic single-task models, without the need for multi-task training.

\subsection*{Data Augmentation}
 When there is no auxiliary data available, data augmentation can also help improve the performance.
Rule-based methods for data augmentation include token-level perturbation \citep{wei-zou-2019-eda}, swapping parts based on dependency trees \citep{sahin-steedman-2018-data}.
\citet{wei-zou-2019-eda} use token-level pertubation such as insertion, deletion and swap.
\citet{sahin-steedman-2018-data} swap parts of sentences base on dependency trees.
\citet{NEURIPS2020_44feb009} propose to use consistency training to leverage unsupervised data using supervised data augmentation methods.
\citet{chen-etal-2020-finding} augment paraphrase identification data by finding paired nodes in a signed graph data.

Other works like \citet{zhang2018mixup,guo-etal-2020-sequence} interpolate between existing examples to augment data in computer vision or NLP.
Oversampling techniques~\citep{10.5555/1622407.1622416, CHARTE2015385, wei-zou-2019-eda} mitigate the class imbalance problem and improve the performance for minority classes.
Language models are also used for data augmentation through back-translation \citep{sennrich-etal-2016-improving}, direct paraphrasing \citep{kumar-etal-2019-submodular}, and replacing tokens by sampling from a language model's distribution \citep{kobayashi-2018-contextual}.
Back-translation \citep{sennrich-etal-2016-improving} and other direct paraphrasing models \citep{kumar-etal-2019-submodular} can augment data by creasing paraphrases for sentences in the original dataset.
They can also be used for data augmentation by replacing tokens by sampling from a language model's distribution \citep{kobayashi-2018-contextual}.

We focus on using auxiliary data to augment primary data.
The above traditional data augmentation methods on a single data source can be used with our method at the same time.

\section{Conclusion}
In this paper, we propose a technique using two-stage training with auxiliary data for attack technique classification.
We demonstrated that adding selected auxiliary data only to rare classes helps alleviate the data imbalance issue for the primary data.
The inclusion improves the classification accuracy on rare classes while maintaining competitive accuracy on common classes.
Further, we pretrained our own BERT model, \ourbert{}, and showcased its effectiveness over existing pretrained security-domain foundation models.

Although we focused on the cyber-attack classification task in this paper, our method is general, model-agnostic, and compatible with any existing training method.
We hope that this work paves the way for researchers to better utilize auxiliary data in other domains through the two-stage training process.




\bibliography{anthology,custom}

\end{document}